\documentclass[lettersize,journal]{IEEEtran}
\usepackage{amsmath,amsfonts}
\usepackage{algorithmic}
\usepackage{array}
\usepackage[caption=false,font=normalsize,labelfont=sf,textfont=sf]{subfig}
\usepackage{textcomp}
\usepackage{stfloats}
\usepackage{url}
\usepackage{verbatim}
\usepackage{graphicx}
\usepackage{multirow}
\usepackage{hyperref}
\def\BibTeX{{\rm B\kern-.05em{\sc i\kern-.025em b}\kern-.08em
    T\kern-.1667em\lower.7ex\hbox{E}\kern-.125emX}}
\usepackage{balance}
\begin{document}
\title{A Persian Benchmark for Joint Intent Detection and Slot Filling}

\author{Masoud Akbari, Amir Hossein Karimi, Tayyebeh Saeedi, Dr. Zeinab Saeidi, Kiana Ghezelbash, Fatemeh Shamsezat, Dr. Mohammad Akbari, Dr. Ali Mohades
\thanks{Masoud Akbari is with the department of Mathematics and Computer Science, Amirkabir University of Technology (email: ma.akbari421@aut.ac.ir)}
\thanks{Amir Hossein Karimi is with the department of Mathematics and Computer Science, Amirkabir University of Technology (email: ahkarimi@aut.ac.ir)}
\thanks{Tayyebeh Saeedi is with the department of Mathematics and Computer Science, Amirkabir University of Technology (email: t.saeedi@aut.ac.ir)}
\thanks{Dr. Zeinab Saeidi is with the department of Mathematics and Computer Science, Amirkabir University of Technology (email: zsaeidi2007@aut.ac.ir)}
\thanks{Kiana Ghezelbash is with the department of Mathematics and Computer Science, Amirkabir University of Technology (email: kghezelbash@aut.ac.ir)}
\thanks{Fatemeh Shamsezat is with the department of Mathematics and Computer Science, Amirkabir University of Technology, and Department of computer science, College of science, Fasa University, 74617-81189, Fasa, Iran. (email: shamsezat@aut.ac.ir)}
\thanks{Dr. Mohammad Akbari is with the department of Mathematics and Computer Science, Amirkabir University of Technology (email: akbari.ma@aut.ac.ir)}
\thanks{Dr. Ali Mohades is with the department of Mathematics and Computer Science, Amirkabir University of Technology (email: mohades@aut.ac.ir)}
}

\markboth{IEEE/ACM Transactions on Audio, Speech, and Language Processing}%
{A Persian Benchmark for Joint Intent Detection and Slot Filling}

\maketitle

\begin{abstract}
  Natural Language Understanding (\textbf{NLU}) is important in today's technology as it enables machines to comprehend and process human language, 
  leading to improved human-computer interactions and advancements in fields such as virtual assistants, chatbots, and language-based AI systems. 
  This paper highlights the significance of advancing the field of \textbf{NLU} for low-resource languages. 
  With intent detection and slot filling being crucial tasks in \textbf{NLU}, the widely used datasets \textbf{ATIS} and \textbf{SNIPS} have been utilized in the past. 
  However, these datasets only cater to the English language and do not support other languages. 
  In this work, we aim to address this gap by creating a Persian benchmark for joint intent detection and slot filling based on the \textbf{ATIS} dataset. 
  To evaluate the effectiveness of our benchmark, we employ state-of-the-art methods for intent detection and slot filling.
\end{abstract}


\section{Introduction}
\IEEEPARstart{N}{atural} Language Understanding (\textbf{NLU}) is an essential component of any goal-oriented dialogue system, 
where it attempts to identify the user's need by processing an utterance. 
NLU aims to automatically identify what the user wants to do (or \textit{intent} of user), 
extract associated arguments (or \textit{slots}), and respond correctly to satisfy the user's request \cite{uATIS}. 
In such a context, intent detection and slot filling are essential components of NLU. 
In the last decade, several works investigate these tasks, however, most of them work on English Language Understanding \cite{intent1, intent_survey, cite_intent}.

Some works on NLU have been done in other languages, such as Romanian \cite{Romanian}, 
Chinese \cite{Chinese}, Arabic \cite{Arabic}, and Vietnamese \cite{Vietnamese}.
Despite the fact that the Persian language is the 24th most spoken language in the world (almost 78 million speakers) \cite{Ethnologue}, 
there is no work on Persian Language Understanding due to the lack of a Persian public dataset.
Without such a dataset, it would be difficult for researchers to work in NLU or, in general, in Natural Language Processing (\textbf{NLP}) tasks.
To address this problem, we develop a dataset in the Persian language and apply some of the state-of-the-art approaches to introduce a public benchmark in the Persian language.

One of the most widely used datasets in intent detection and slot filling is the \textbf{ATIS} (Air Transform Information System) dataset. 
This dataset is originally in English and includes intent and slot annotations. 
In 2018, Upadhyay et al. extended the ATIS to Hindi and Turkish \cite{ATIS_E}. 
Also, recently the authors in \cite{MultiATIS++} introduced the \textbf{MultiATIS++} and extended ATIS to six more languages in four families. 
Hence MultiATIS++ covers nine languages including English, Spanish, German, French, Portuguese, Chinese, Japanese, Hindi, and Turkish. 
As mentioned before, the MultiATIS++ covers four families, which are Indo-European, Sino-Tibetan, Japonic, and Altaic \cite{MultiATIS++}. 
However, the MultiATIS++ fails to cover the Persian language, which belongs to the Indo-European family.

In this paper, we want to introduce a Persian public benchmark for all researchers in the NLP field. 
We use original ATIS dataset and apply a semi-automatic approach to provide a Persian dataset. 
To provide a baseline for researchers, 
we perform a comparison time analysis to have a comprehensive and public Persian benchmark. 
This benchmark will help researchers who are interested in Persian Language Understanding task to improve their work. 

Our contributions are summarized as follows:
\begin{itemize}
    \item We construct the first intent detection and slot filling dataset in Persian by extending ATIS dataset to Persian,
    \item We perform a comparative analysis of state-of-the-art models on the Persian dataset which we constructed,
    \item We make our benchmark available to the public for research and educational purposes. Our benchmark, we believe, will serve as a starting point for future Persian NLU research and applications.
\end{itemize}

\section{Problem statement}
\noindent In this paper, we focus on intent detection and slot filling, which are two main tasks in NLU.
We apply the algorithms that can perform intent detection and slot filling simultaneously on our provided Persian dataset.
At first, we explain these problems and their applications in NLU.

\subsection{Intent detection}
\noindent As mentioned above, intent detection aims to find out what the user wants to do and, based on that, respond appropriately to the user. 
So, usually, the intent detection task is modelled as a sentence classification problem. 
Since the sentences in the intent detection task are coming from spoken language utterances, 
the intent classification has some differences from other classification tasks. 
Hence, before applying any classification algorithms in order to classify the intent, we should extract the features which have semantic information in the sentence. 
Generally, research around intent detection in NLU is presented in four main areas: search engines, dialogue systems, question answering systems, and text categorization \cite{intent_survey}.

\subsection{Slot filling}

\noindent Slot filling is typically treated as sequence labelling problem in which contiguous sequences of words are tagged with sementic labels.
The input of slot filling task is the sentence consisting of a sequence of words, and the output is a sequence of tags one for each word \cite{intent_survey,9033997,6737243,6998838}.

Dialogue systems are the most common area of NLP that uses slot filling. 
The role of slot filling in dialogue systems is to interpret natural language sentences of users, 
using which the system can determine what information to retrieve or what task to complete for the user. 
Online shopping websites with task-oriented dialog systems are also integrated with slot filling models. 
Also, with the use of slot filling, it is possible to understand product search queries better and provide shopping assistance to customers \cite{intent_survey}.

\begin{figure*}[ht!]
  \centering
  \includegraphics[scale=0.25]{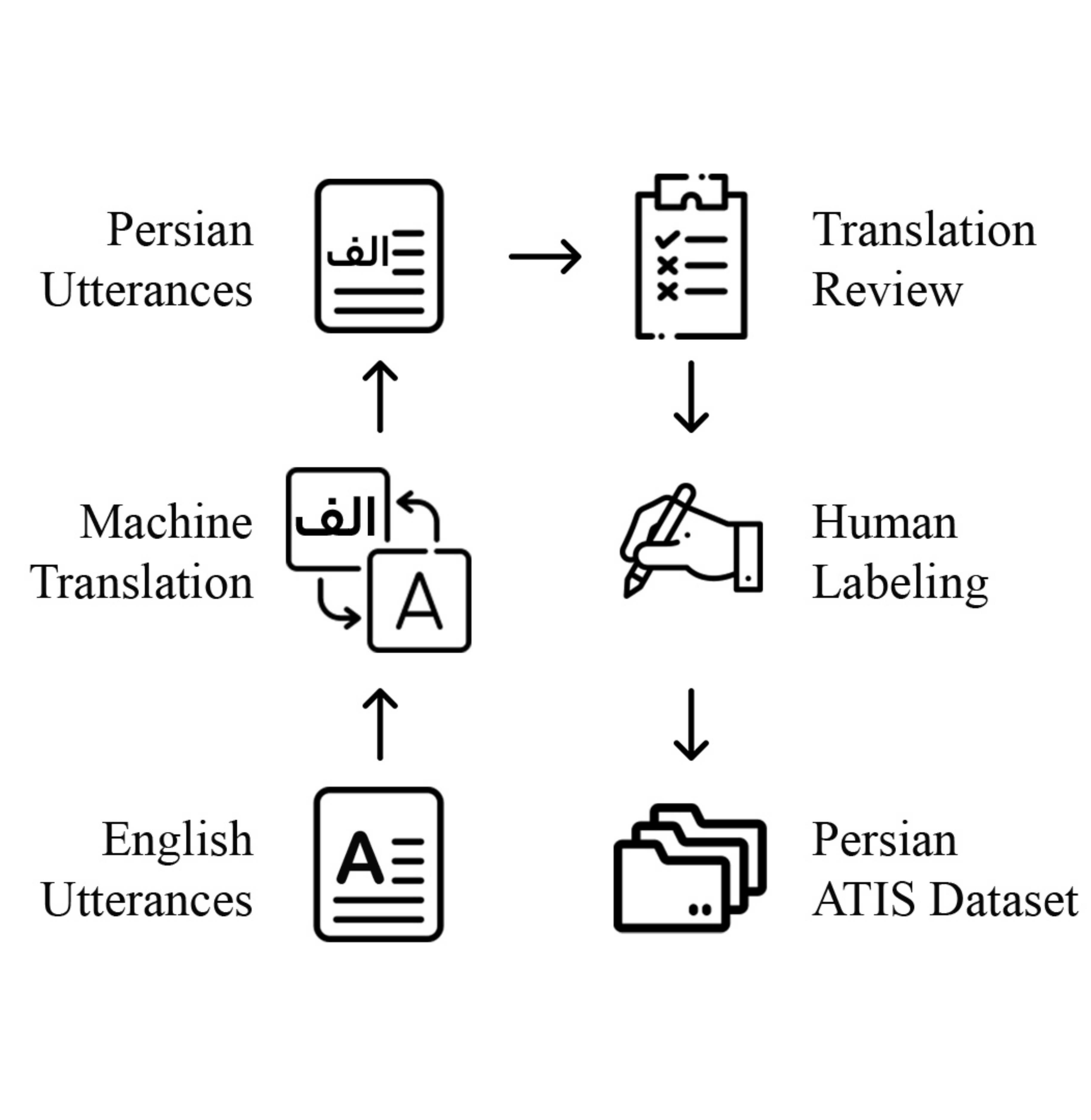}
  \caption{\footnotesize{Process flow of constructing Persian ATIS via a semi-automatic approach using translation}}
  \label{fig1}
\end{figure*}

\section{Description of persian dataset}
\noindent There are Persian datasets for NLP tasks like question-answering \cite{pquad, pecoq, persianquad}, 
language modeling \cite{PWiki}, or sentiment analysis \cite{sharami2020deepsentipers}. 
However, there is no Persian benchmark dataset for the NLU task. 
As NLU has a significant effect on the performance of a chatbot, having a Persian dataset is necessary to do more researches in this branch. 
ATIS is an English benchmark dataset for NLU.
As mentioned before, in this work, we chose the ATIS dataset and translated that into the Persian language automatically. 
We chose this dataset because equivalences of ATIS utterances in Persian are more meaningful than datasets like SNIPS. 
For example, the sentence "{\it{Find The Passion of Michel Foucault novel.}}" contains the name of an object that is not common in Persian. 
On the other side, the translation of a sentence in the ATIS dataset, like "{\it{I want to fly from Boston at 8 am and arrive in Denver at 11 in the morning,}}"
contains more common words in Persian. 
Furthermore, the ATIS was constructed on an actual popular travel domain, making it meaningful in various languages.

The ATIS dataset comprises 5871 utterances in English (4978 utterances for training and 893 for evaluation). 
Fig.~\ref{fig1} shows the process flow of creating the Persian ATIS dataset.
The first step was to translate ATIS to Persian with a machine translation. 
We used the \textbf{EasyNMT}\footnote{https://github.com/UKPLab/EasyNMT} package as the machine translator. 
EasyNMT contains three different models for neural machine translation and covers more than 100 languages. 
mBart50 \cite{mbart50} was one of the models in EasyNMT that covers the Persian language, too. 
We also used mBart50 to translate English ATIS to Persian.

\begin{figure}[h!]
  \centering
  \includegraphics[scale=0.5]{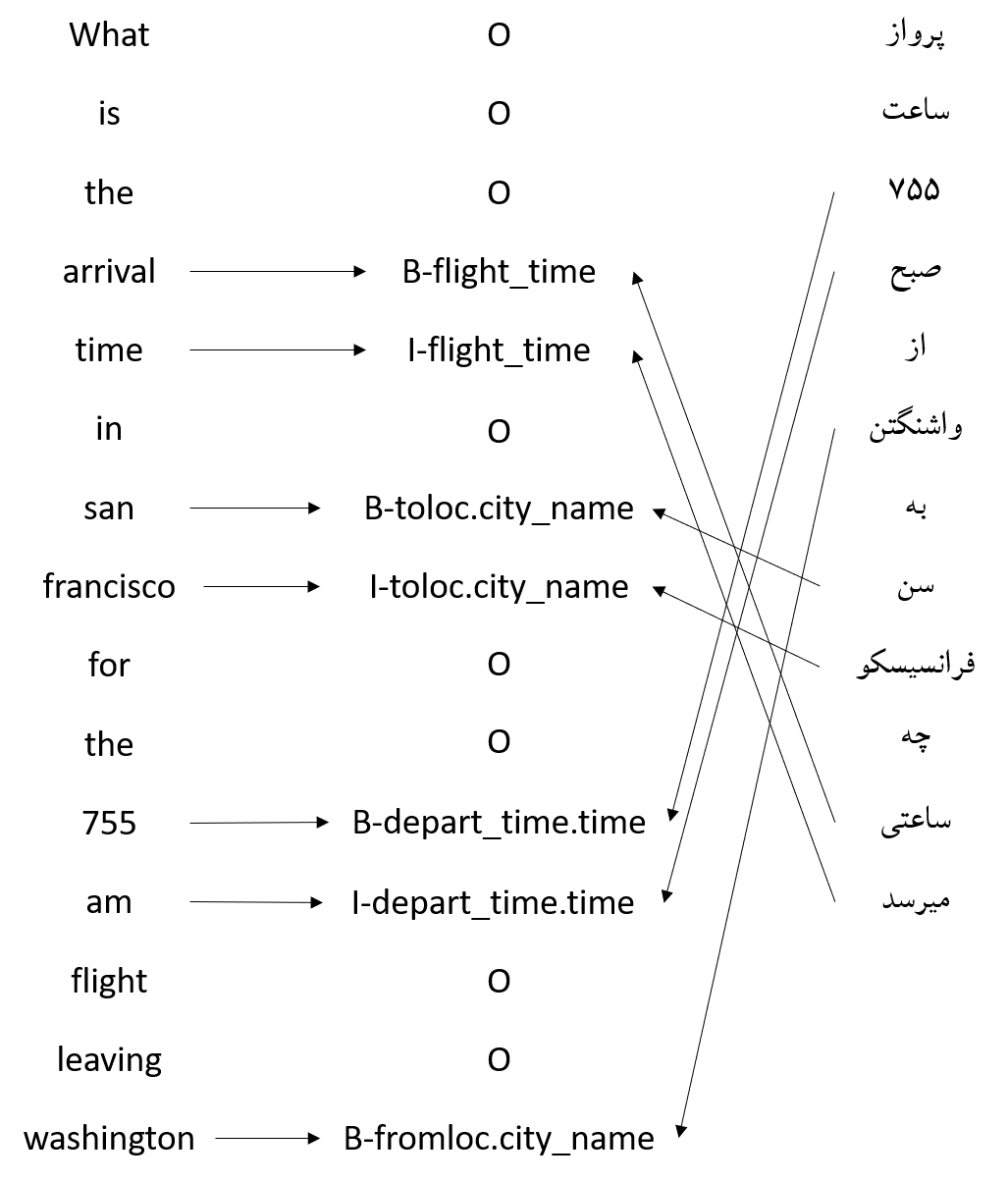}
  \caption{An example of supplants of words after translation and the need to relabel slot tags.}
  \label{fig2}
\end{figure}

Since the machine translations may make minor mistakes, 
such as misplacing translated words in the destination language, 
all the translations were double-checked and corrected by Persian natives. 
The third step was labeling. 
The intent of an utterance remains the same during translation, 
so no effort is needed to set intent labels of utterances. 
However, as some words are supplanted after translation, 
it is necessary to do the slot labeling process. 
An example of this problem is shown in Fig.~\ref{fig2}.
It should be noted that NLP experts did the labeling process.

As well as other natural language processing tasks, preprocessing the input text is necessary in the last step. 
There is a challenge in Persian NLP datasets related to the similarity between the Arabic and Persian alphabets. 
It means that two words with the same shape may be considered as different tokens because of the different ASCI codes of their characters. 
Normalizing the data and unifying the characters can solve this problem. 
We used the \textbf{Hazm}\footnote{https://github.com/roshan-research/hazm} Persian NLP library with some other changes for normalization. 
Useless punctuation also has been removed. The number of unique characters changed from 113 to 69. 
There were some exceptional names in the translated data, 
such as the ATIS dataset city or airport names, 
even after the translation process, which we substituted with Persian equivalents.

\section{A taxonomy of intent detection and slot filling}

\begin{figure*}[h!]
    \centering
    \includegraphics[scale=0.4]{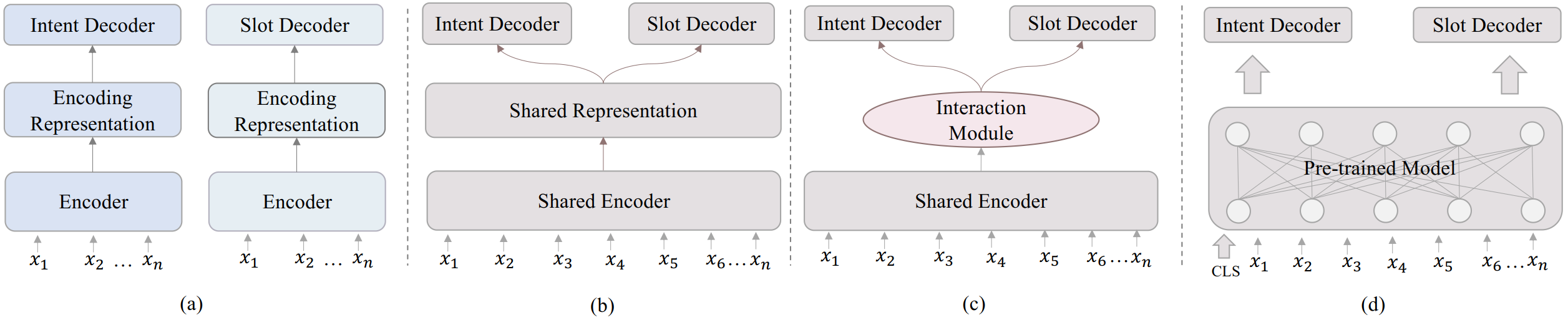}
    \caption{(a) Single models. (b) Implicit Joint Modeling. (c) Explicit Joint Modeling. (d) Pre-trained Model Paradigm. \cite{qin2021survey}}
    \label{fig3}
\end{figure*}

\noindent Some state-of-the-art methods were chosen to evaluate the Persian ATIS dataset. 
These methods had to cover various approaches to produce a reliable benchmark. 
We chose these methods based on the taxonomy defined in \cite{qin2021survey}. 
This taxonomy classifies methods in NLU into single models, joint models, and pre-trained paradigms. 
The paradigms are visualized in fig. \ref{fig3} which was presented initially in \cite{qin2021survey}.

In the single models paradigm, there is no interaction between intent detection and slot-filling tasks. 
Hence, intent detection and slot-filling will not be able to take advantage of the correlation between them and share helpful information to rich better performance. 
On the other side, joint models leverage the shared knowledge between intent detection and slot-filling. 

The joint paradigm is also separated into two classes: explicit and implicit. 
The implicit joint modeling only uses a shared encoder (representation) and does not explicitly model the interaction, which may result in low interpretability and performance. 
The explicit joint modeling paradigm is classified into a single and bidirectional flow interaction. The single flow only considers a single information flow from intent to the slot. 
In contrast, the bidirectional flow means that model considered the cross-impact between intent detection and slot filling.

Lastly, the pre-trained paradigm is defined as a paradigm where pre-trained language models (like BERT) are used as encoders to extract representation. Pre-trained models can provide rich semantic features, which can help to improve the performance of SLU tasks.

In the next section, we will discuss the selected methods based on the taxonomy and provide their performance on the Persian ATIS dataset.

\begin{table*}[h!]
  \centering
  \caption{Intent accuracy percent of state-of-the-art models on English and Persian ATIS dataset}
  \label{intent}
  \begin{tabular}{|c|c|c|c|}
  \hline
  \textbf{Taxonomy}               & \textbf{Model}              & \textbf{English} & \textbf{Persian}  \\ \hline
  \textbf{Single}                 & \textbf{CNN-LSTM-CRF \cite{CNN}}       & 93.62            & 89.70             \\ \hline
  \multirow{8}{*}{\textbf{Joint}} & \textbf{CNN-LSTM-CRF \cite{CNN}}       & 93.73            & 91.83              \\ \cline{2-4} 
                                  & \textbf{Attention RNN \cite{JointAtt}}      & 93.84            & 90.93          \\ \cline{2-4} 
                                  & \textbf{Slot-Gated \cite{SGated}}         & 94.62            & 92.94             \\ \cline{2-4} 
                                  & \textbf{SF-ID,SF-first \cite{BiDir}}     & 96.65   & 92.38   \\ \cline{2-4} 
                                  & \textbf{SF-ID+CRF,SF-first \cite{BiDir}} & 97.31            & 92.83                \\ \cline{2-4} 
                                  & \textbf{SF-ID,ID-first \cite{BiDir}}     & 97.09            & 92.83                 \\ \cline{2-4} 
                                  & \textbf{SF-ID+CRF,ID-first \cite{BiDir}} & 95.41            & 92.05                  \\ \cline{2-4}
                                  & \textbf{Co-Interactive transformer (Glove) \cite{coint}} & \textbf{97.54} &    \underline{93.73}    \\ \hline
  \textbf{Pre-trained} & \textbf{JointBERT \cite{chen2019bert}} & \underline{97.42} & \textbf{97.65}       \\ \hline
  \end{tabular}
\end{table*}
\begin{table*}[h!]
  \centering
  \caption{Slot F1-score percent of state-of-the-art models on English and Persian ATIS dataset}
  \label{slot}
  \begin{tabular}{|c|c|c|c|}
  \hline
  \textbf{Taxonomy}               & \textbf{Model}              & \textbf{English} & \textbf{Persian} \\ \hline
  \textbf{Single}                 & \textbf{CNN-LSTM-CRF \cite{CNN}}       & 94.46            & \underline{88.41}            \\ \hline
  \multirow{8}{*}{\textbf{Joint}} & \textbf{CNN-LSTM-CRF \cite{CNN}}       & 85.31            & 81.68            \\ \cline{2-4} 
                                  & \textbf{Attention RNN \cite{JointAtt}}      & \underline{95.59}            & 87.96   \\ \cline{2-4} 
                                  & \textbf{Slot-Gated \cite{SGated}}         & 94.91            & 87.70            \\ \cline{2-4} 
                                  & \textbf{SF-ID,SF-first \cite{BiDir}}     & 94.65            & 85.38            \\ \cline{2-4} 
                                  & \textbf{SF-ID+CRF,SF-first \cite{BiDir}} & 94.72            & 85.56            \\ \cline{2-4} 
                                  & \textbf{SF-ID,ID-first \cite{BiDir}}     & 95.06            & 85.32            \\ \cline{2-4} 
                                  & \textbf{SF-ID+CRF,ID-first \cite{BiDir}} & 94.55            & 85.57            \\ \cline{2-4}
                                  & \textbf{Co-Interactive transformer (Glove) \cite{coint}} & \textbf{95.69} & 86.14 \\ \hline
  \textbf{Pre-trained} & \textbf{JointBERT \cite{chen2019bert}} & 95.20 & \textbf{96.75} \\ \hline
  \end{tabular}
\end{table*}
\begin{figure*}[h!]
  \centering
  \includegraphics[scale=0.4]{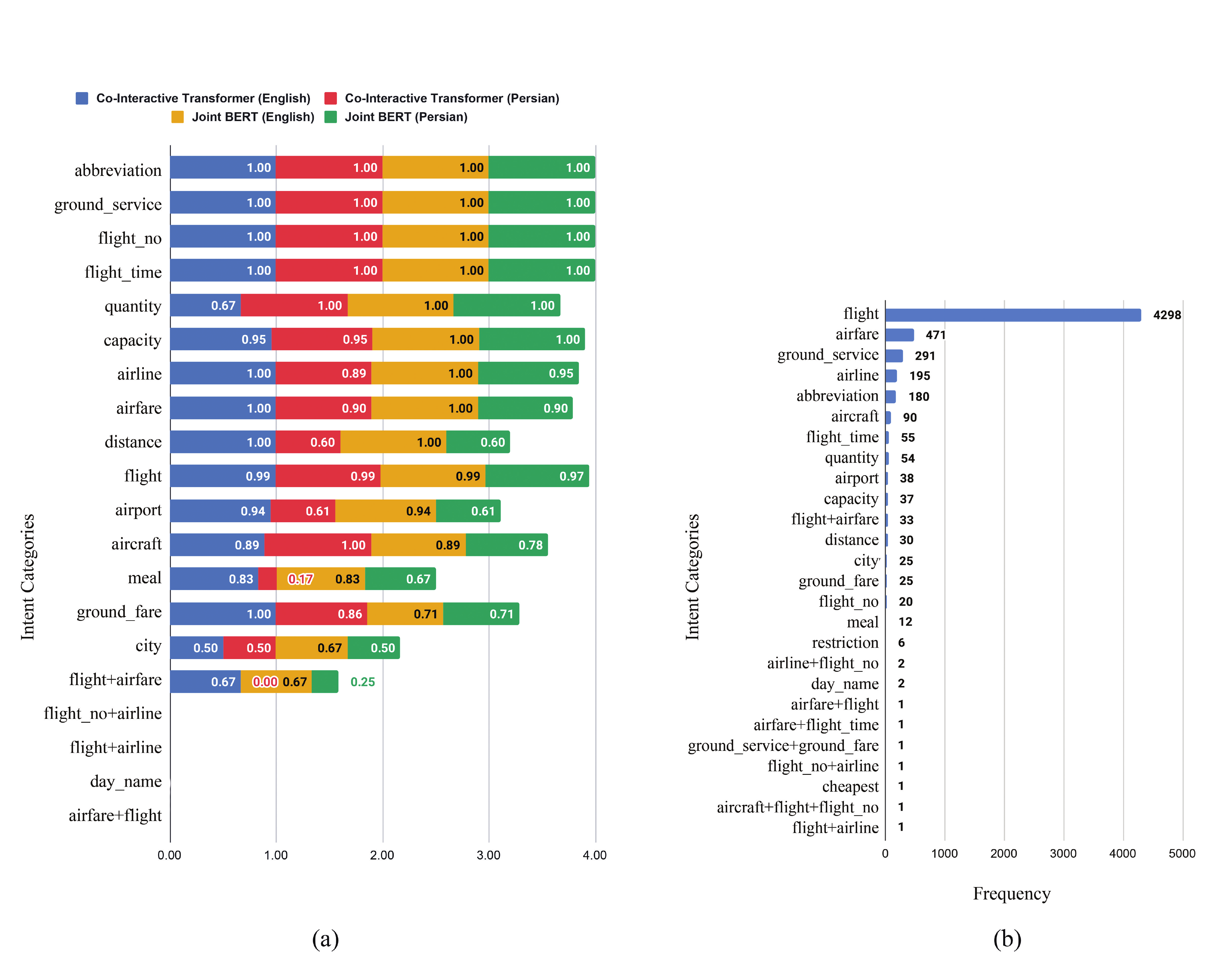}
  \caption{(a) Comparison of the best methods in intent detection on both Persian and English datasets; 
          each bar is a stack composed of the accuracy of each method, and the length of each part is related to its performance. 
          (b) All existing intent categories in the dataset and their frequency}
  \label{fig4}
\end{figure*}
\begin{figure*}[h!]
  \centering
  \includegraphics[scale=0.29]{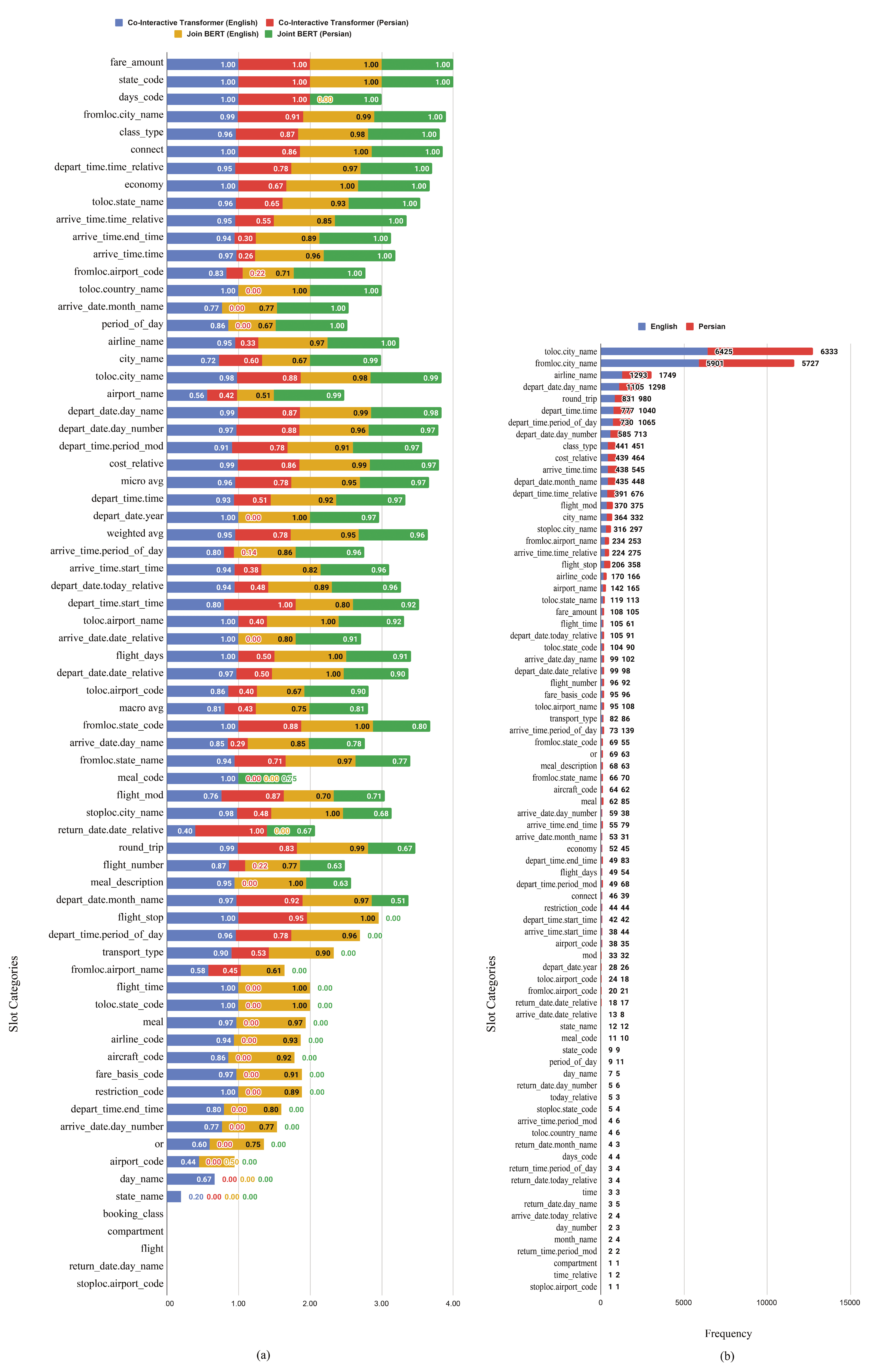}
  \caption{(a) Comparison of the best methods in slot filling on both Persian and English datasets; 
          each bar is a stack composed of the F1-score of each method, and the length of each part is related to its performance. 
          (b) All existing slot categories in the dataset and their frequency; 
          since the frequency of each slot differs in each dataset, frequencies are shown separately.}
  \label{fig5}
\end{figure*}

\section{Experiments}
\noindent In this section, we evaluate and compare state-of-the-art Intent detection and Slot filling models on our proposed corpus and the English ATIS. We apply following models based on taxonomy on dataset:

\subsection{Models}
\noindent The following models are selected based on the taxonomy discussed in the previous section. 

\begin{itemize}
  \item \textbf{SF-ID \cite{BiDir}}: This model lies in the explicit joint modeling paradigm. 
                          In this approach, A bidirectional model considers the cross-impact of slot filling and intent detection. 
                          The first part is calculating the context vectors by an attention mechanism. 
                          Next is the network consisting of slot-filling and intent detection subnets, 
                          where their order can be customized (slot-filling first or intent detection first). 
                          Finally, a CRF layer is added above the slot-filling subnet outputs to consider the correlations between the labels in neighborhoods. 
                          It jointly decodes the best chain of labels of the utterance.
  \item \textbf{Attention RNN \cite{JointAtt}}: This approach uses a neural network based on the attention mechanism to solve the problem of joint intent detection and slot filling. 
                                      This approach also lies in the implicit joint modeling category. 
                                      The proposed model consists of an encoder and a decoder, where the encoder uses a BiLSTM network and the decoder uses an LSTM. 
                                      The concatenation of the forward and backward walks related to the last hidden state will be used to initiate the decoder's hidden state. 
                                      The context vector is also used to compute the attention. 
                                      The output of the decoder will be used for slot filling, and the output of the encoder for intent detection.
  \item \textbf{CNN-LSTM-CRF \cite{CNN}}: This model can be trained as a single or joint model. 
                                The proposed method in this paper first uses a CNN network to encode utterances at the word level and capture morphological information, such as prefixes and suffixes. 
                                The outputs of the CNN will go through a BiLSTM network, and the outputs of BiLSTM are passed to a CRF layer for slot filling and a classifier to detect the intent.
  \item \textbf{Slot-Gated \cite{SGated}}: The Attention-based RNN model achieves the best performance on joint ID and SF (where the attention weights of ID and SF are independent). 
                                This explicitly modeled method proposes a slot-gated structure, which focuses on learning the relationship between intent and slot attention vector, 
                                and obtains a better semantic frame through global optimization. 
                                They explicitly model the relationships between slots and intents using a single flow interaction from intents to slots. 
                                The proposed slot gating model introduces an additional gate that utilizes intent context vectors to model slot-intent relations to improve slot-filling performance. 
                                First, the slot context vector and intent context vector combine. 
                                In other words, in the encoder, for each time stamp, a slot context vector and a global intent context vector are computed using an attention mechanism. 
                                They introduced slot-gate g, computed as a function of context vectors. 
                                Then, g is used as a weight for the input's slot label of the i'th word. 
                                Then to obtain both slots and intent labels, they formulated the objective as the conditional probability of the understanding result (slot filling and intent prediction) given the input word sequence.
\item \textbf{Co-Interactive transformer \cite{coint}}: As a multi-task learning model, the CIT model can fill slots and detect intent simultaneously. 
                                                      This network consists of two sub-networks: the slot-filling network and the intent detection network. 
                                                      A co-interactive mechanism connects these two sub-networks, allowing them to interact and share information. 
                                                      Slot-filling networks encode input text using transformers and predict slot labels based on linear layers. 
                                                      Another transformer-based encoder encodes the input text, and a linear layer predicts the intent label for the entire text. 
                                                      A multi-head attention mechanism is used to implement the co-interactive mechanism, 
                                                      which allows the slot-filling and intent detection networks to interact and share information. 
                                                      Specifically, the multi-head attention mechanism calculates the attention weights between the slot-filling and intent detection representations and 
                                                      then combines the two representations according to these attention weights. 
                                                      This method is an explicit joint modeling method. 
                                                      However, since BERT representations can be applied along with this model, 
                                                      the proposed method can also be classified as a pre-trained modeling paradigm. 
                                                      In this paper, we only used it as an explicit joint modeling method as the official implementation using BERT was unavailable.
\item \textbf{JointBERT \cite{chen2019bert}}: This method uses representations from the pre-trained BERT model to solve the joint intent detection and slot-filling task. 
                                            A softmax activation function is used over the representation computed for the CLS tag of each input to detect intents and a 
                                            softmax over the final hidden states of other tokens to classify slots. This method lies in the pre-trained paradigm.
                                            We used the BERT's multi-lingual version to apply the model to Persian and English ATIS \cite{DBLP:journals/corr/abs-1810-04805}.
                                                      
\end{itemize}

\subsection{Evaluation Metrics}
\noindent The two metrics used for comparison are accuracy for intent detection and F1-score for slot-filling. 
As the ATIS dataset is imbalanced for both intent and slot classes, 
we have also computed accuracy and F1-score for each class separately to measure how well the selected models perform.

\subsection{Implementation Setup}
\noindent At first, we split the Persian and English ATIS into the train, test, and validation sets. 
The splitting was done in such a way as to keep the correspondence the same between English and Persian data. 
The validation split is a randomly selected 20\% of the primary training set. 
Since the original ATIS is imbalanced based on intent classes and slot labels, the test set, which we split, has 20 intent categories out of 26 intent classes. 
In addition, the test set has 69 slot categories out of 83 slot labels. 
The categories of intent and slot with frequency in the dataset are shown in Fig.~\ref{fig4} and Fig.~\ref{fig5}, respectively. 
Plus, the experiments' configuration for each model is the same as the original setup. 
Moreover, we used the available codes provided by the authors.

\subsection{Comparison Performance on the English and Persian ATIS datasets}

\subsubsection{Intent Detection}

The accuracy for intent detection is shown in Table~\ref{intent}. 
As can be seen, the Co-Interactive transformer method has the best performance for the English and the second-best performance for the Persian dataset. 
In contrast, the JointBERT method has the best performance for the Persian and the second-best performance for the English dataset. 
However, the difference in the performance of the Co-Interactive Transformer and JointBERT is negligible for the English dataset. 
One of the factors that help JointBERT outperform other methods is the quality of representations given by the pre-trained BERT model. 
Generally, the performance of methods for the Persian dataset is lower, which is also related to the representations and embeddings. 
For example, the Persian version of Glove is 50-dimensional, which may lead to less separable representations than the 300-dimensional English Glove embeddings.


Fig.~\ref{fig4} shows the ability of the models with the best accuracy on the English and the Persian dataset to detect each class separately for the test dataset. 
The frequency of each intent in the dataset is also shown. 
In most cases, the accuracy decreases as the number of samples for a class decreases. 
The reason is that the model learned to detect more common classes.

\subsubsection{Slot Filling}

The results of slot filling are provided in Table 2. 
The methods with the best performance are the same as those in intent detection - the Co-Interactive Transformer for English and JointBERT for Persian. 
The Attention RNN method and the JointBERT have the second and third-best performances, with a poor difference. 
On the other side, CNN-LSTM-CRF in the single learning mode has the second-best performance for Persian, with a vast difference. 
The reasons for the difference in the performance in English and Persian are the same as we noted previously for the intent detection section.

Fig.~\ref{fig5} compares the performance of the best methods in slot filling on both Persian and English datasets for each label. 
The frequency of each slot in Persian differs from that in English. 
The difference results from the changes in the number of parts of words during translation. 
Also, the relation between the performance of the methods on each slot with the frequency of that slot is the same as the relation between detecting intents and the frequency of each intent.

\section{Acknowledgement}
We want to express our great appreciation to Shahrzad Zolghadr, Fatemeh Ghoochani, Fatemeh Inanloo, and Sepehr Moradi, who helped and gave us their time to facilitate reaching the goals of this research.

\section{Conclusion}
\noindent Research into natural Persian language processing has gotten more challenging due to the size of the number of individuals who speak Persian, 
as well as the lack of datasets devoted to intent recognition and slot filling in Persian. 
In order to address this issue, we presented the first publicly available benchmark in the Persian language for intent detection and slot filling. 
We then investigated the state-of-the-art models for intent detection and slot filling to apply them to our recently founded benchmark and explain how they perform on this particular dataset. 

We hope that the newly developed Persian benchmark for joint intent detection and slot filling will provide valuable training data as well as a novel and demanding testing ground for NLU models.

\end{document}